\pdfoutput=1

\documentclass[11pt]{article}

\usepackage[preprint]{acl}

\usepackage{times}
\usepackage{latexsym}

\usepackage[T1]{fontenc}

\usepackage[utf8]{inputenc}

\usepackage{microtype}

\usepackage{inconsolata}

\usepackage{graphicx}

\title{ParetoRAG: Leveraging Sentence-Context Attention for \\ Robust and Efficient Retrieval-Augmented Generation}

\author{
  \textbf{Ruobing Yao\textsuperscript{1,2}},
  \textbf{Yifei Zhang\textsuperscript{3}\thanks{Corresponding author: zhanyi.zyf@alibaba-inc.com}},
  \textbf{Shuang Song\textsuperscript{1,2}},
  \textbf{Yuhan Liu\textsuperscript{1,2}},
  \textbf{Neng Gao\textsuperscript{1}\thanks{Corresponding author: gaoneng@iie.ac.cn}},
  \textbf{Chenyang Tu\textsuperscript{1}}
  \\
  \textsuperscript{1}Institute of Information Engineering, Chinese Academy of Sciences, Beijing, China,
  \\
  \textsuperscript{2}School of Cybersecurity, University of Chinese Academy of Sciences, Beijing, China
  \\
  \textsuperscript{3}Alibaba Group, Beijing, China,
}
\usepackage{microtype}
\usepackage{graphicx}
\usepackage{booktabs} 
\usepackage{svg}
\usepackage{graphicx}   
\usepackage{subcaption} 
\usepackage{placeins} 
\usepackage{float}

\usepackage[utf8]{inputenc}
\usepackage[titletoc,title]{appendix}
\usepackage[tikz]{mdframed}
\usepackage{multirow}
\usepackage[normalem]{ulem}
\useunder{\uline}{\ul}{}

\usepackage{amsmath}
\usepackage{amssymb}
\usepackage{mathtools}
\usepackage{amsthm}
\usepackage{graphicx}
\usepackage{adjustbox}

\usepackage{enumitem}
\usepackage{url}

\usepackage{colortbl}
\usepackage{xcolor}

\newcommand{\redtext}[1]{\colorbox{red!30}{#1}}
\newcommand{\bluetext}[1]{\colorbox{blue!30}{#1}}

\newcommand{\poscell}[2]{\cellcolor{blue!#1}{#2}} 
\newcommand{\negcell}[2]{\cellcolor{red!#1}{#2}}   

\begin{document}
\maketitle
\begin{abstract}

While Retrieval-Augmented Generation (RAG) systems enhance Large Language Models (LLMs) by incorporating external knowledge, they still face persistent challenges in retrieval inefficiency and the inability of LLMs to filter out irrelevant information.
We present \textbf{ParetoRAG}, an unsupervised framework that optimizes RAG systems through sentence-level refinement guided by the \textbf{Pareto principle}. 
By decomposing paragraphs into sentences and dynamically re-weighting core content while preserving contextual coherence, ParetoRAG achieves dual improvements in both retrieval precision and generation quality without requiring additional training or API resources. This framework has been empirically validated across various datasets, LLMs, and retrievers. 
Futhermore, we show that ParetoRAG's architectural improvements are orthogonally compatible with adaptive noise-robust models, enabling retrieval-augmented optimization and robust training to mutually enhance generation quality. This highlights architectural refinements and noise mitigation as complementary, offering insights for integrating retrieval augmentation with robustness enhancement.
\end{abstract}
\section{Introduction}

\begin{figure}[htbp!]
    \centering
    \includegraphics[width=0.95\linewidth]{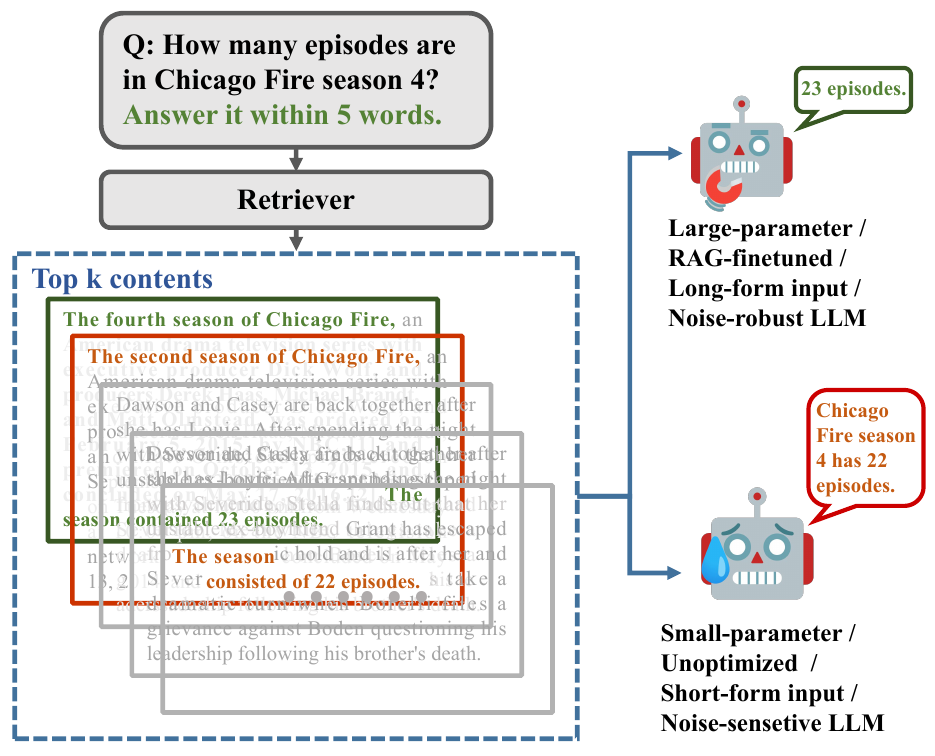}
    \caption{The examples show that a large amount of noise impedes the LLM from acquiring accurate knowledge from the retrieved content and could potentially misdirect its reasoning. Finding the correct answer relies on the ability of LLM to identify a small portion of key information.}
    \label{fig1:redundency phenomenon}
\end{figure}

With the development of Large Language Models (LLMs), their general capabilities have become increasingly powerful \cite{achiamGPT4TechnicalReport2023, dubey2024llama}. However, even the most advanced LLMs still face challenges with factual errors \cite{minFActScoreFinegrainedAtomic2023, huangFactAlignLongformFactuality2024}. One major limitation lies in their static parametric memory, which prevents them from adapting to dynamically evolving knowledge demands or covering unknown domains beyond their training data \cite{kasaiRealTimeQAWhats2023}. Furthermore, LLMs are prone to generating hallucinations that appear plausible but lack factual accuracy \cite{huangSurveyHallucinationLarge2024}. These challenges significantly hinder the performance of LLMs in knowledge-intensive tasks \cite{ramInContextRetrievalAugmentedLanguage2023}. To address these limitations, Retrieval-Augmented Generation (RAG) \cite{lewisRetrievalaugmentedGenerationKnowledgeintensive2020, xiongApproximateNearestNeighbor2020, izacardUnsupervisedDenseInformation2021} integrates relevant passages from external databases into the input context, effectively enhancing the reliability and performance of models in open-domain question answering and dynamic knowledge retrieval tasks. 


\begin{figure*}[t]
    \centering
    \includegraphics[width=1\linewidth]{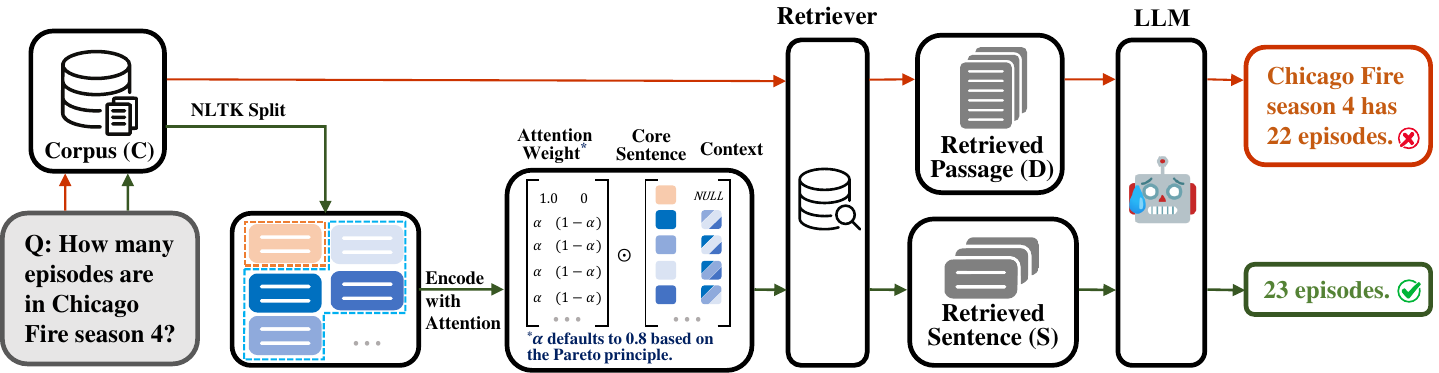}
    \caption{Comparison of the traditional RAG (red path) and ParetoRAG(green path). The traditional method retrieves and directly uses entire passages, which often introduces redundant information, leading to inaccurate answers. In contrast, our method utilizes a preprocessed sentence-level corpus, assigning higher weights to key sentences while appropriately preserving and weighting contextual information to avoid loss of coherence. Inspired by the Pareto principle (the 80/20 rule), this design emphasizes critical information while maintaining necessary semantic consistency. The selected sentences are then fed into the LLM, resulting in more accurate answers.}
    \label{fig2:ParetoRAG pipeline}
\end{figure*}

However, the effectiveness of RAG highly depend on the quality of the retrieved information \cite{fan2024survey}. Additionally, interference from redundant information and increased input length are critical factors that significantly impact model performance.
In the retrieval stage, the relevance scores of core sentences can be overshadowed by those of redundant content at the same passage level, reducing the prominence of key information in the retrieved content. In the generation stage, retrieving excessive content to provide rich context can result in overly lengthy inputs, which may cause the model to lose focus and diminish its ability to concentrate on critical information \cite{jin2024longcontextllmsmeetrag, largelanguagemodelsCanbeEasilyDistractedbyInrelevantContext}. As Figure \ref{fig1:redundency phenomenon} shows that core sentences account for only a small portion of the top k retrieved content. Excessive irrelevant or redundant information hinders the ability of the model to extract accurate knowledge and increases the risk of generating hallucinations \cite{zhang2024raft, liu-etal-2024-lost}.
In addition, in the zero-shot Chain-of-Thought \cite{wei2022chain} prompting setup, the ability of LLM to follow instructions shows a significant decline as the input size increases. The model tends to directly generate answers before completing reasoning steps, and this tendency becomes more pronounced as inputs grow longer \cite{levy-etal-2024-task}. 


Retrieval-augmented language models (RALMs) \cite{zhang2024raft, lin2024radit}, Long-context LLMs \cite{dubey2024llama, team2024gemini}, and Adaptive Noise-Robust Model \cite{yoran2024making, fangEnhancingNoiseRobustness2024a} can be considered as solutions. These models enhance the ability to process long-form text and improve the robustness to noisy information, enabling them to focus more effectively on key information and reduce the impact of redundant content. However, these approaches generally require additional training resources and high computational costs for further training and model fine-tuning. Another possible solution is to reduce the granularity of retrieval from the document level to the sentence level \cite{leePhraseRetrievalLearns2021, chenDenseRetrievalWhat2024a}. However, this approach may inadvertently lose some important contextual information (e.g., in the example in Figure \ref{fig1:redundency phenomenon}, "The season" refers to Chicago Fire season 4 in one context and season 2 in another), which is crucial for accurately answering the given query \cite{choiDecontextualizationMakingSentences2021}. Therefore, we propose a method that does not require additional training resources while effectively preserving contextual information and reducing document redundancy.

In this work, we present \textbf{ParetoRAG}, an unsupervised framework built upon the RAG system. Our approach leverages a preprocessed sentence-level corpus, assigning higher weights to key sentences while carefully preserving and weighting contextual information to maintain coherence. Drawing inspiration from the \textbf{Pareto principle} (the 80/20 rule), ParetoRAG prioritizes critical information while ensuring semantic consistency, effectively enhancing both the retrieval and generation stages of the RAG pipeline. Notably, ParetoRAG requires neither additional training resources nor extra API
calls. The overall ParetoRAG framework is illustrated in Figure \ref{fig2:ParetoRAG pipeline}. 

We validate ParetoRAG across three datasets over three retrievers. 
Our main contributions are as follows:
\begin{itemize}
    \item A plug-and-play method named ParetoRAG is proposed to achieve the decomposition from paragraph-level to sentence-level, effectively retaining contextual information during the retrieval stage without requiring additional training.
    \item ParetoRAG achieves notable improvements in accuracy and fluency while reducing token consumption to approximately 30\% of the original cost. Furthermore, it demonstrates strong generalization, as this conclusion is consistently validated across various datasets, LLMs, and retrievers.
    \item We investigate the methodological compatibility between ParetoRAG's improvements and the adaptive noise-robust model. The findings suggest that retrieval-augmented architectures and robust training strategies can be orthogonally beneficial, providing architectural-level enhancements that complement rather than interfere with existing noise mitigation approaches.
\end{itemize}


\section{Related Work}






\textbf{Retrieval-Augmented Generation with Noisy Context} RAG \cite{guuREALMRetrievalaugmentedLanguage2020, lewisRetrievalaugmentedGenerationKnowledgeintensive2020} is considered a useful method to address hallucinations, which improves the input questions of generative LLM with retrieved documents. It usually provides an extra knowledge source from a specific corpus, i.e., Wikipedia, which greatly improves the performance of LLM in a variety of tasks, especially in the knowledge-intensive ones \cite{ramInContextRetrievalAugmentedLanguage2023}. However, due to the limitation of retrieval capabilities, retrieval-augmented systems inevitably introduce irrelevant or partially relevant knowledge to the models \cite{yinALCUNALargeLanguage2023}. In recent years, the impact of noisy information on the performance of RAG systems has received increasing attention \cite{zhuFreeLBEnhancedAdversarial2019, yuChainofNoteEnhancingRobustness2024}.  Some studies \cite{jiaAdversarialExamplesEvaluating2017, creswellSelectionInferenceExploitingLarge2022} have shown that the introduction of irrelevant noise significantly degrades model performance. Further analyses \cite{chenBenchmarkingLargeLanguage2025} indicate that as the proportion of noise in the retrieval context increases, the performance of large language models (LLMs) deteriorates significantly. In addition, research \cite{fangEnhancingNoiseRobustness2024a} has explored the effects of different types of noise on RAG systems and found that counterfactual retrieval noise has the most detrimental impact on retrieval systems.

\begin{figure}[ht!]
    \centering
    \includegraphics[width=1\linewidth]{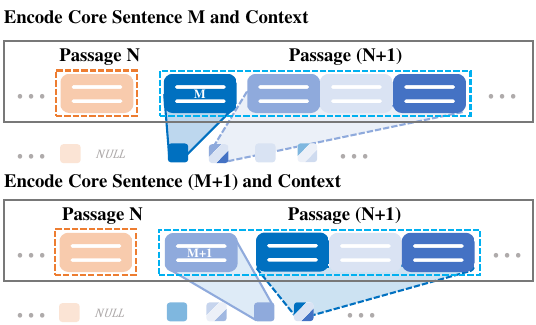}
    \caption{The example of ParetoRAG encodes core sentence M and core sentence (M+1). The content within the same dashed box is split from the same passage. The context of a core sentence consists of sentences from the same passage, excluding the core sentence itself.}
    \label{fig:Pareto Encode Example}
\end{figure} 

\textbf{Advanced RAG} Many advanced approaches have been developed from the original RAG in recent years \cite{kimSuReSummarizingRetrievals2023, zhang2024raft, liuRAISFLearningAnswer2024, patilGorillaLargeLanguage2024}. Considering that retrieval is sometimes unnecessary for some queries, conversely, responses without retrieval are even more accurate in many situations. SelfRAG \cite{asaiSelfRAGLearningRetrieve2023a} is proposed to selectively retrieve knowledge and introduce a critic model to decide whether to retrieve. SAIL \cite{luoSearchAugmentedInstruction2023} is tuned on instructions to insert the retrieved documents before the instructions. RECOMP \cite{xu2024recomp} is designed to refine the retrieved passages by either abstractively or extractively summarizing them with additional models.

Compared with recent studies \cite{hwangDSLRDocumentRefinement2024, chenDenseRetrievalWhat2024a} that are the most relevant to our work, a main difference should be highlighted. Dense X reduces information redundancy by degrading documents into proposition sentences. To decompose paragraphs into propositions, the authors trained a fine-tuned text generation model to supplement contextual information. In contrast, our approach utilizes the Sentence-Context Weighted Attention mechanism to supplement contextual information without requiring any fine-tuning.

\section{Method}
In this section, we introduce a novel framework, ParetoRAG, for improving the accuracy of retrieval results by leveraging sentence-level weighting inspired by the Pareto Principle, integrated into the RAG system. Notably, ParetoRAG requires neither additional training resources nor extra API calls.

\subsection{Encoding Step}
The encoding step involves extracting the core sentence and its corresponding context from the passages and encoding them into dense vector representations. This process ensures the inclusion of both the key information from the core sentence and the supplementary information from its surrounding context, enabling semantically rich representation for subsequent retrieval.

\textbf{Passage Segmentation. }The passages in the retrieval corpus $\mathcal{C}$ are segmented into sentences using NLTK. The retrieval corpus is represented as a collection of passages:
\[
\mathcal{C} = \{P_1, P_2, \dots, P_m\},
\]
where $m$ is the total number of passages. Each passage $P_j \in \mathcal{C}$ is represented as a sequence of sentences:
\[
P_j = \{s_1^j, s_2^j, \dots, s_{n_j}^j\},
\]
where $s_i^j$ represents the $i$-th sentence in the $j$-th passage, and $n_j$ is the total number of sentences in $P_j$.

\textbf{Core Sentence and Context Extraction. }For each passage $P_j \in \mathcal{C}$, every sentence $s_i^j$ is considered a core sentence, and its corresponding context is defined as the concatenation of all other sentences in the same passage, excluding $s_i^j$ itself. Formally, for a passage $P_j = \{s_1^j, s_2^j, \dots, s_{n_j}^j\}$, the context for the core sentence $s_i^j$ is defined as:
\\[\bigskipamount]
    \resizebox{0.5\textwidth}{!}{
    $
    \text{Context}(s_i^j) = 
    \begin{cases} 
    \{s_1^j, \dots, s_{i-1}^j, s_{i+1}^j, \dots, s_{n_j}^j\}, & \text{if } n_j > 1, \\
    \text{NULL}, & \text{if } n_j = 1.
    \end{cases}
    $
    }
\\[\bigskipamount]
Here, $\text{Context}(s_i^j)$ captures the surrounding information in the passage $P_j$ without including the core sentence $s_i^j$ itself. This ensures that each core sentence $s_i^j$ can be analyzed independently while still being informed by its contextual sentences. If a passage consists of only one sentence ($n_j = 1$), the context is defined as $\text{NULL}$.

\textbf{Encode Core Sentence, Context and Query. }For each core sentence $s_i^j$ and its corresponding context $\text{Context}(s_i^j)$, a configurable encoder $\text{Enc}_\theta(\cdot)$ is applied to obtain their vector representations. Here, $\theta$ represents the model selection parameter, which determines the specific encoder to be used (e.g., Contriever, ANCE, or DPR). 


The core sentence $s_i^j$, its context $\text{Context}(s_i^j)$, and the query are encoded into $d$-dimensional vector representations using the same encoder $\text{Enc}_\theta(\cdot)$. The encoding process is as follows:
\[
\begin{aligned}
\mathbf{h}_{\text{core}}^i &= \text{Enc}_\theta(s_i^j), \\
\mathbf{h}_{\text{context}}^i &= \text{Enc}_\theta(\text{Context}(s_i^j)), \\
\mathbf{q} &= \text{Enc}_\theta(\text{Query}),
\end{aligned}
\]
where $\mathbf{h}_{\text{core}}^i$, $\mathbf{h}_{\text{context}}^i$, and $\mathbf{q}$ are all $d$-dimensional vectors. These representations are used for similarity computation and ranking.

These vector representations $\mathbf{h}_{\text{core}}^i$, $\mathbf{h}_{\text{context}}^i$, and $\mathbf{q}$ are then used to calculate the similarity and rank in subsequent steps. Figure \ref{fig:Pareto Encode Example} shows the example of ParetoRAG encodes core sentence M and core sentence (M+1).

\subsection{Retrieval Step}
The retriever step takes the encoded core sentence, context, and query vectors to compute their similarity and rank the results for retrieval. This process consists of the following key substeps:

\textbf{Sentence-Context Weight Adjustment} 
To balance the contributions of the core sentence and its context, an attention mechanism assigns weights based on a hyperparameter $\alpha$. The weighted representation is computed as: 
\\[\bigskipamount]
    \resizebox{0.5\textwidth}{!}{
    $
    \mathbf{h}_{\text{weighted}}^i =
    \begin{cases}
    \mathbf{h}_{\text{core}}^i, & \text{if } \text{Context}(s_i^j) = \text{NULL}, \\
    \alpha \cdot \mathbf{h}_{\text{core}}^i + (1 - \alpha) \cdot \mathbf{h}_{\text{context}}^i, & \text{otherwise}.
    \end{cases}
    $
    }
\\[\bigskipamount]
This mechanism ensures that both the key information from the core sentence and the supplementary information from its context are considered during similarity computation.

\textbf{Similarity Computation}
Following previous studies \cite{lewisRetrievalaugmentedGenerationKnowledgeintensive2020, zouPoisonedRAGKnowledgeCorruption2024}, the similarity between the weighted sentence representation $\mathbf{h}_{\text{weighted}}^i$ and the query vector $\mathbf{q}$ is computed using dot similarity by default.
    \[
    \text{Sim}(s_i^j, \mathbf{q}) = \mathbf{h}_{\text{weighted}}^i \cdot \mathbf{q}.
    \]
This step quantifies how closely each sentence-context pair matches the semantic meaning of the query.

\textbf{Top-$k$ Sentence Ranking} 
The top-$k$ sentences are ranked based on their similarity scores in descending order:
\[
\text{Top-}k = \operatorname{arg\,top}_k \left( \text{Sim}(s_i^j, \mathbf{q}) \right),
\]
where $\operatorname{arg\,top}_k$ returns the indices of the $k$ sentences with the highest similarity scores. These top-$k$ sentences are selected as the retrieval results, providing the most relevant information based on the query.

\begin{table*}[ht!]
\resizebox{\textwidth}{!}{ 
\begin{tabular}{lccccccccccccccc}
\hline
\multicolumn{1}{c|}{\multirow{2}{*}{Model}} & \multicolumn{4}{c}{NQ(acc)} & \multicolumn{4}{c|}{Hotpot(acc)} & \multicolumn{4}{c|}{MS(mauve)} & \multicolumn{3}{c}{MS(rouge)} \\
\multicolumn{1}{c|}{} & \# tok & Contriever & ANCE & DPR & \# tok & Contriever & ANCE & \multicolumn{1}{c|}{DPR} & \# tok & Contriever & ANCE & \multicolumn{1}{c|}{DPR} & Contriever & ANCE & DPR \\ \hline
\multicolumn{16}{c}{Without RAG} \\ \hline
Vicuna-7B & \multirow{4}{*}{/} & \multicolumn{3}{c}{23.2} & \multirow{4}{*}{/} & \multicolumn{3}{c}{16.1} & \multirow{4}{*}{/} & \multicolumn{3}{c}{88.3} & \multicolumn{3}{c}{40.5} \\
Vicuna-13B &  & \multicolumn{3}{c}{28.2} &  & \multicolumn{3}{c}{20.2} &  & \multicolumn{3}{c}{82.1} & \multicolumn{3}{c}{40.7} \\
Llama2-7B-chat &  & \multicolumn{3}{c}{20.9} &  & \multicolumn{3}{c}{16.0} &  & \multicolumn{3}{c}{85.6} & \multicolumn{3}{c}{36.2} \\
Llama2-13B-chat &  & \multicolumn{3}{c}{29.9} &  & \multicolumn{3}{c}{18.4} &  & \multicolumn{3}{c}{90.1} & \multicolumn{3}{c}{39.6} \\ \hline
\multicolumn{16}{c}{Naive RAG} \\ \hline
Vicuna-7B & \multirow{4}{*}{895} & 33.2 & 36.1 & 41.9 & \multirow{4}{*}{757} & 25.0 & 22.3 & 23.9 & \multirow{4}{*}{576} & 84.1 & 84.9 & 87.9 & 35.8 & 35.6 & 37.5 \\
Vicuna-13B &  & 37.4 & 41.0 & 45.6 &  & 22.6 & 20.2 & 22.7 &  & 86.8 & 87.7 & 87.0 & 37.8 & 36.9 & 36.9 \\
Llama2-7B-chat &  & 33.2 & 37.9 & 40.8 &  & 23.6 & 23.4 & 23.3 &  & 85.0 & 88.6 & 89.2 & 33.6 & 34.4 & 35.4 \\
Llama2-13B-chat &  & 38.3 & 39.6 & 42.7 &  & 27.1 & 25.3 & 26.8 &  & 77.5 & 87.0 & 88.8 & 33.2 & 33.5 & 34.7 \\ \hline
\multicolumn{16}{c}{Recomp \cite{xu2024recomp}} \\ \hline
Vicuna-7B 
& \multirow{4}{*}{26} & \poscell{8}{35.9} & \poscell{9}{39.3} & \poscell{4}{43.4} 
& \multirow{4}{*}{41} & \poscell{16}{29.0} & \poscell{14}{25.5} & \poscell{17}{27.9} 
& \multirow{4}{*}{26} & \negcell{6}{79.2} & \negcell{1}{83.8} & \negcell{3}{85.3} 
& \poscell{13}{40.4} & \poscell{19}{42.3} & \poscell{10}{41.2} \\ 

Vicuna-13B 
&  & \negcell{2}{36.8} & \poscell{1}{41.5} & \negcell{7}{42.5} 
&  & \poscell{28}{28.9} & \poscell{25}{25.4} & \poscell{13}{25.6} 
&  & \negcell{2}{85.4} & \negcell{2}{85.7} & \poscell{0}{87.2} 
& \poscell{7}{40.5} & \poscell{13}{41.6} & \poscell{9}{40.3} \\ 

Llama2-7B-chat 
&  & \negcell{27}{24.3} & \negcell{17}{31.5} & \negcell{17}{33.9} 
&  & \poscell{13}{26.7} & \negcell{4}{22.4} & \poscell{8}{25.2} 
&  & \negcell{70}{25.7} & \negcell{54}{41.2} & \negcell{57}{38.5} 
& \poscell{9}{36.7} & \poscell{14}{39.2} & \poscell{6}{37.6} \\ 

Llama2-13B-chat 
&  & \negcell{18}{31.5} & \negcell{9}{36.2} & \negcell{7}{39.5} 
&  & \poscell{11}{\ul{30.1}} & \negcell{0}{25.2} & \poscell{8}{28.9} 
&  & \negcell{48}{40.4} & \negcell{47}{45.9} & \negcell{53}{41.6} 
& \poscell{12}{37.0} & \poscell{17}{39.0} & \poscell{9}{37.7} \\ \hline

\multicolumn{16}{c}{ParetoRAG (Ours)} \\ \hline
Vicuna-7B 
& \multirow{4}{*}{232} & \poscell{10}{36.6} & \poscell{20}{43.4} & \poscell{11}{46.7} 
& \multirow{4}{*}{229} & \negcell{2}{25.4} & \poscell{13}{25.3} & \negcell{3}{24.7} 
& \multirow{4}{*}{183} & \poscell{5}{88.1} & \poscell{3}{87.4} & \poscell{4}{91.5} 
& \poscell{19}{42.5} & \poscell{25}{\ul{44.4}} & \poscell{15}{43.4} \\ 

Vicuna-13B 
&  & \poscell{5}{39.1} & \poscell{8}{44.2} & \poscell{6}{\ul{48.2}} 
&  & \poscell{18}{26.7} & \poscell{28}{25.9} & \poscell{15}{26.0} 
&  & \negcell{3}{84.3} & \poscell{1}{88.8} & \negcell{1}{86.7} 
& \poscell{10}{41.6} & \poscell{18}{43.6} & \poscell{15}{42.6} \\ 

Llama2-7B-chat 
&  & \poscell{2}{34.0} & \poscell{10}{41.7} & \poscell{4}{42.3} 
&  & \poscell{4}{24.6} & \poscell{7}{25.0} & \poscell{3}{24.0} 
&  & \poscell{8}{92.0} & \poscell{1}{89.9} & \poscell{4}{92.6} 
& \poscell{20}{40.4} & \poscell{26}{43.3} & \poscell{18}{41.9} \\ 

Llama2-13B-chat 
&  & \negcell{6}{36.1} & \poscell{6}{41.8} & \poscell{11}{47.4} 
&  & \negcell{4}{25.9} & \poscell{3}{26.1} & \negcell{6}{25.2} 
&  & \poscell{19}{\ul{92.2}} & \poscell{4}{90.1} & \poscell{4}{92.0} 
& \poscell{17}{39.0} & \poscell{26}{42.2} & \poscell{18}{40.8} \\ \hline
\end{tabular}
}
\caption{Overall experiment results of three retrievers on three tasks, based on top 30 recall contents. The deeper the \redtext{red background}, the lower the relative improvement. In contrast, the deeper the \bluetext{blue background}, the higher the relative improvement compared to Naive RAG. The \underline{underlined numbers} indicate the best-performing results on the current dataset.}
\label{table: main results}
\end{table*}

\subsection{Generation Step}
After the ranking step, the top-$k$ ranked sentences, denoted as $\text{Top-}k$, are passed to the language model $M$ to generate the final answer $a$ for the given query $\mathbf{q}$. The generation process integrates the query and the retrieved sentences to produce a response that is both accurate and contextually relevant. The generation step can be formalized as:
\[
a = \text{Generate}(\mathbf{q}, \text{Top-}k; M),
\]
where $\text{Generate}(\cdot)$ represents the generation function that combines the query $\mathbf{q}$, the retrieved top-$k$ sentences $\text{Top-}k$, and the language model $M$ to produce the output $a$.



\section{Experiment Setups}

In this section, we describe the experimental setup for evaluating ParetoRAG across various scenarios. The specific model parameters can be found in Appendix \ref{appendix: Statistics of models}. The selection and meaning of the evaluation metrics can be found in Appendix \ref{appendix: Evaluation Metrics}.

\subsection{Datasets.} 

We experiment on three different open-domain QA datasets as the retrieval source: Natural Questions (NQ) \cite{kwiatkowskiNaturalQuestionsBenchmark2019}, HotpotQA \cite{yangHotpotQADatasetDiverse2018}, and MS-MARCO \cite{nguyen2016ms}, where each dataset has a knowledge database. 
These datasets encompass different tasks, such as open-domain question answering, multi-hop reasoning, and long-form answer generation. 
Each dataset also contains a set of questions. We randomly sampled 1,000 data paris for testing. Table \ref{table:statistic-dataset} shows statistics of text unit counts before and after ParetoRAG encoding.

\subsection{Dense Retrieval Models}
We compare the performance of the three following unsupervised, semi-supervised,  or supervised dense retriever models. Following previous studies~\cite{lewisRetrievalaugmentedGenerationKnowledgeintensive2020a}, by default, we use the dot product between the embedding vectors of a question and a text in the knowledge database to calculate their similarity score. 

\textbf{Contriever} \cite{izacardUnsupervisedDenseInformation2021} is an unsupervised retriever implemented using a BERT-base encoder. Contriever is contrastively trained on segment pairs constructed from unlabeled documents in Wikipedia and web crawl data.

\textbf{ANCE} \cite{xiongApproximateNearestNeighbor2020} is a dual-encoder BERT-base model designed for dense retrieval tasks. It is trained using weakly supervised signals from query-document pair labels, typically sourced from datasets such as MS-MARCO.

\textbf{DPR} \cite{karpukhinDensePassageRetrieval2020} is a dual-encoder  BERT-base model fine-tuned on passage retrieval tasks directly using the question-passage pair labels from NQ, TQA \cite{joshiTriviaQALargeScale2017}, SQuAD \cite{rajpurkarSQuAD100000Questions2016} and WebQ \cite{berantSemanticParsingFreebase2013}.

\begin{figure*}[ht!]
    \centering
    \begin{subfigure}[b]{0.45\textwidth} 
        \centering
        \begin{subfigure}[b]{0.48\textwidth} 
            \centering
            \includegraphics[width=\linewidth]{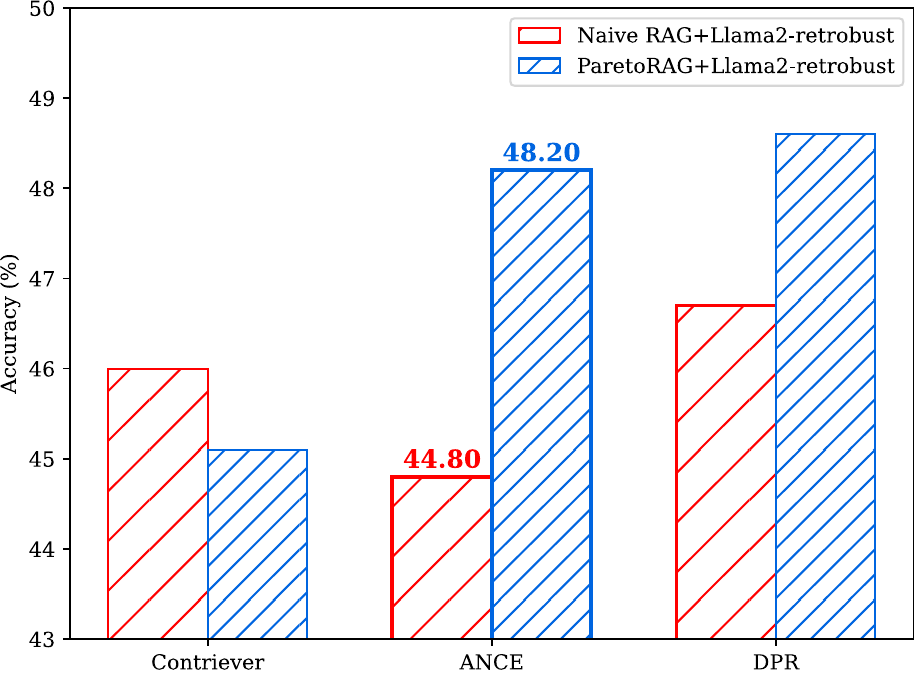} 
            \caption{NQ (Top 10)}
            \label{fig:PareroRAGforRobustLLM-a}
        \end{subfigure}
        \begin{subfigure}[b]{0.48\textwidth}
            \centering
            \includegraphics[width=\linewidth]{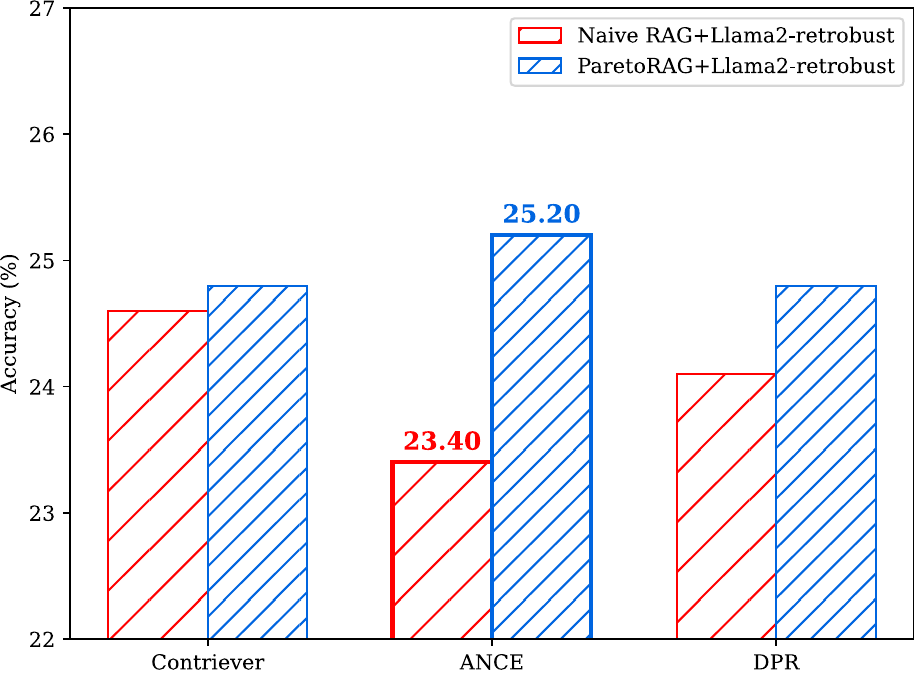} 
            \caption{HotpotQA (Top 10)}
            \label{fig:PareroRAGforRobustLLM-b}
        \end{subfigure}
    \end{subfigure}
    \begin{subfigure}[b]{0.45\textwidth} 
        \centering
        \begin{subfigure}[b]{0.48\textwidth}
            \centering
            \includegraphics[width=\linewidth]{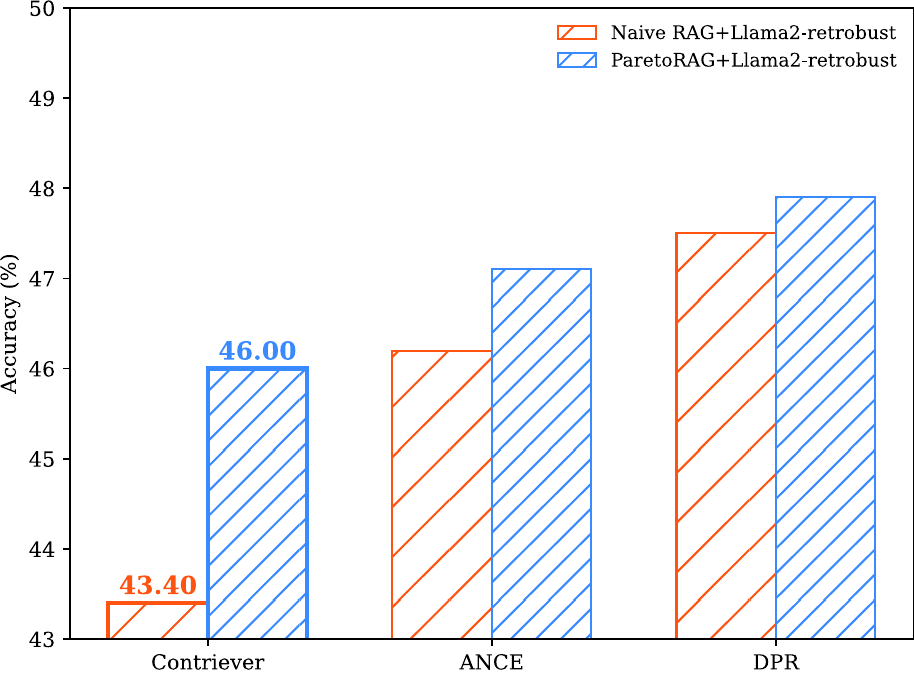} 
            \caption{NQ (400 words)}
            \label{fig:PareroRAGforRobustLLM-c}
        \end{subfigure}
        \begin{subfigure}[b]{0.48\textwidth}
            \centering
            \includegraphics[width=\linewidth]{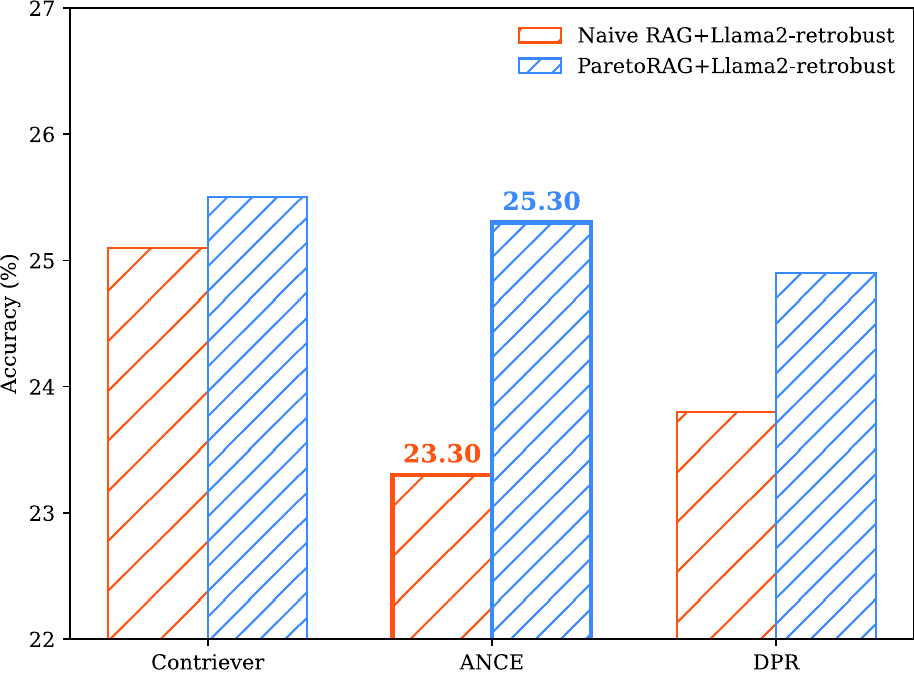} 
            \caption{HotpotQA (400 words)}
            \label{fig:PareroRAGforRobustLLM-d}
        \end{subfigure}
    \end{subfigure}
\caption{Comparison of ParetoRAG and Naive RAG on the adaptive noise-robust LLM (\texttt{llama-2-13b-peft-nq-retrobust} and \texttt{llama}-\texttt{2-13b-peft-hotpotqa-retrobust} \cite{yoran2024making}): (a)(b) show performance under the same recall size (Top 10), while (c)(d) illustrate performance under the same input word count (400).}
\label{fig:ParetoRAGforRobustLLM}
\end{figure*}

\subsection{Baselines}

For these three baselines, we evaluated publicly available instruction-tuned models, such as Vicuna-7B and Vicuna-13B \cite{zhengJudgingLLMasaJudgeMTBench2023} , as well as models trained and reinforced with private data, including Llama2-7B-Chat and Llama2-13B-Chat \cite{touvronLlama2Open2023}. 

\textbf{Baselines without retrievals. }We evaluate the performance of various LLMs without employing RAG technology across multiple datasets.

\textbf{Baselines with Naive RAG.} We employ the most basic RAG technique, without incorporating complex retrieval optimization methods or advanced generation mechanisms, relying solely on the fundamental retrieval-generation workflow.

\textbf{Baselines with SOTA methods.} We implement two advanced approaches: (1) Recomp \cite{xu2024recomp}  using abstractive summarization (excluding extractive variants) to synthesize retrieved passages with dedicated models. (2) LLM trained with adversarial noise to improve robustness \cite{yoran2024making}.

\section{Experimental Results and Analysis}
In this section, we show the overall experimental results with in-depth analyses of our framework.

\subsection{Main Results}

Table \ref{table: main results} presents the results of three retrievers on three datastes, based on top-30 recall contents. Figure \ref{fig:ParetoRAGforRobustLLM} illustrates the performance of ParetoRAG on \path{llama2-13b-retrobust}. From these results, we can conclude the following findings:

\textbf{ParetoRAG, while consuming only about 30\% of the original token cost, still delivers  notable  improvements in accuracy and fluency.} Specifically,  as shown in Table \ref{table: main results}, in NQ, the accuracy of Vicuna-7B + ANCE increases from 36.1\% to 43.4\% (+7.3\%), with the token count reduced to 26\% of the original. Similarly, the accuracy of Llama2-13B-Chat + DPR increases from 42.7\% to 47.4\% (+4.7\%), with the token count reduced to approximately 30\% of the original. In HotpotQA, the accuracy of Vicuna-13B + ANCE improves from 20.2\% to 25.9\% (+5.7\%), with the token count reduced to approximately 30\% of the original. 

In addition, in MS-Marco, ParetoRAG achieves notable improvements in both mauve (fluency) and rouge (correctness) metrics. For example, the fluency score of Llama2-13B-Chat + Contriever increases from 77.5 to 92.2 (+14.7\%), while the token count is reduced to approximately 32\% of the original. Similarly, the fluency score of Vicuna-7B + DPR improves from 87.9 to 91.5 (+3.6\%). In terms of correctness (rouge), the rouge score of Vicuna-13B + ANCE increases from 46.8 to 55.2 (+8.4\%), while Llama2-13B-Chat + ANCE improves from 46.1 to 55.1 (+9\%). These results further highlight ParetoRAG's capability to deliver consistent and measurable improvements in both fluency and accuracy, even with significantly reduced token consumption. 

%

\begin{figure*}[ht!]
    \centering
    \begin{subfigure}[b]{0.31\textwidth} 
        \centering
        \includegraphics[width=\linewidth]{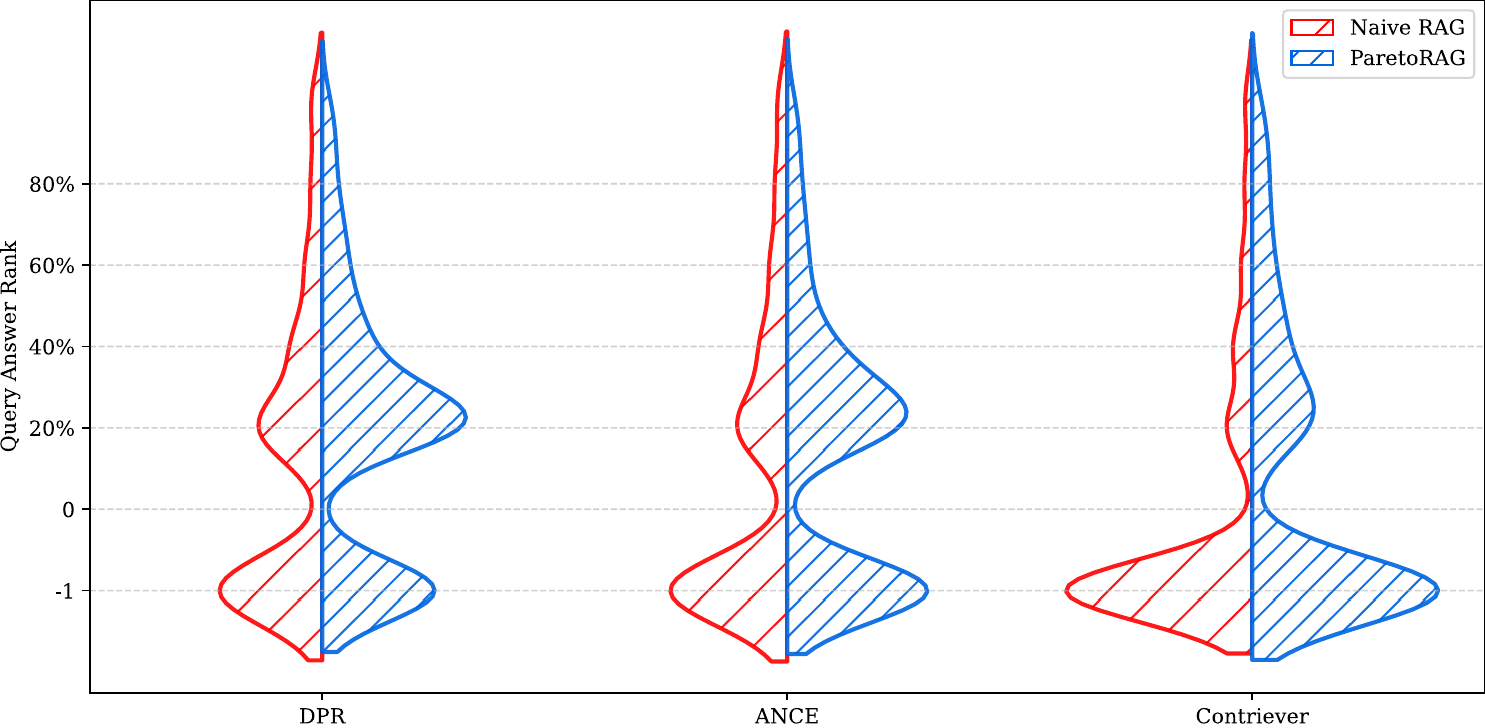}
        \caption{NQ distribution}
        \label{fig:nq distribution}
        \vspace{-5pt} 
    \end{subfigure}
    \begin{subfigure}[b]{0.31\textwidth} 
        \centering
       \includegraphics[width=\linewidth]{{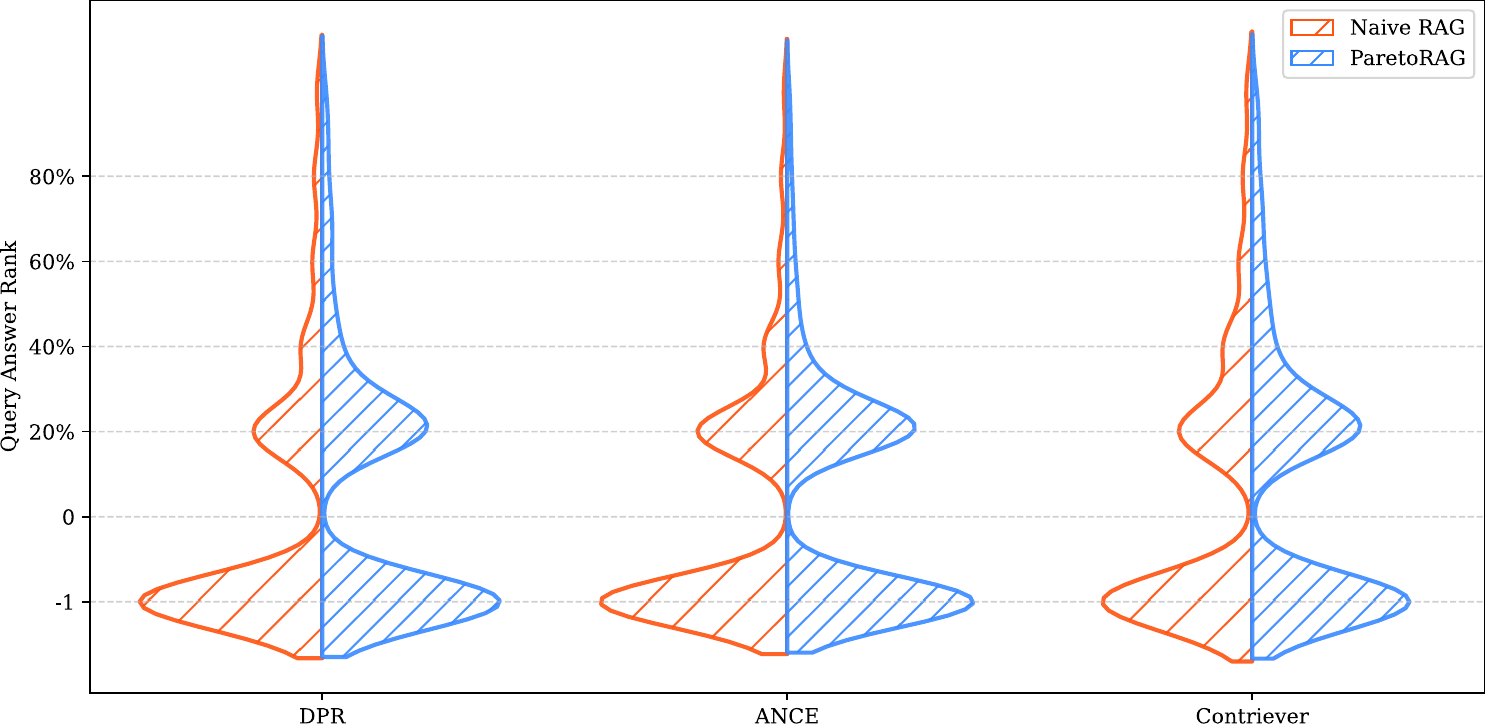}}
        \caption{HotpotQA distribution}
        \label{fig:hotpotqa distribution}
        \vspace{-5pt} 
    \end{subfigure}
    \begin{subfigure}[b]{0.31\textwidth} 
        \centering
        \includegraphics[width=\linewidth]{{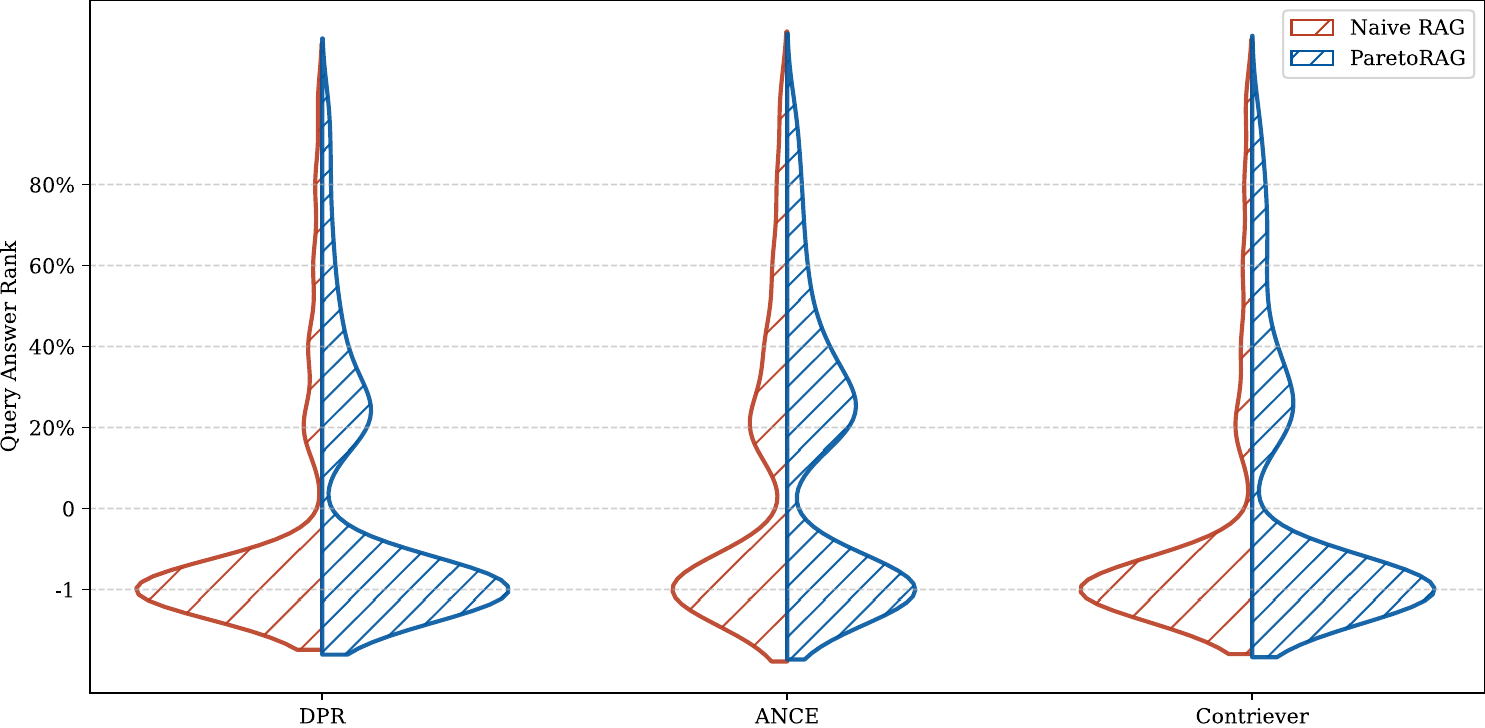}}
        \caption{MS-MARCO distribution}
        \label{fig:msmarco distribution}
        \vspace{-5pt} 
    \end{subfigure}
    \caption{Correct answer rank distributions across different datasets under the the same input word count (400).}
    \label{fig:data-distribution}
\end{figure*}

\textbf{ParetoRAG  demonstrates strong generalizations.} We analyze its effectiveness from three perspectives:

\textit{Effectiveness across multiple datasets: }ParetoRAG consistently improves performance across a diverse range of datasets, including NQ, HotpotQA, and MS-Marco. 
In contrast, Recomp does not have a specialized abstract model for the MS-MARCO task, resulting in a significant drop in performance on MAUVE (fluency) and a smaller improvement on ROUGE-L (correctness) compared to ParetoRAG.

\textit{Compatibility with different types of retrievers:} ParetoRAG proves effective with various retriever types, including Contriever, ANCE, and DPR. This shows that the method is not tied to a specific retriever and adapt well to different retrieval methods. Specific analysis of the impact on retrievers can be found in \ref{sec:impact-of-retriever}.
    
\textit{Applicability across multiple LLMs:} ParetoRAG achieves improvements when applied to large language models, such as Vicuna-7B, Vicuna-13B, Llama2-7B-Chat, and Llama2-13B-Chat. Notably, we also test the method on models trained with anti-noise techniques. As shown in Figure \ref{fig:ParetoRAGforRobustLLM}, the results still show improvements. More detailed analysis can be found in \ref{sec: detailed-analysis-on-robust-llm}.

\subsection{Ablation Study}
We study the impact of core sentence weight, retriever types, and top k size on ParetoRAG. The variation of core sentence weight on HotpotQA and MS-MARCO can be found in Appendix \ref{appendix: Impact of core sentence weight}, while the impact of model parameters on ParetoRAG is detailed in Appendix \ref{appendix: Impact of Model Size on Accuracy with Varying Top K in ParetoRAG}.

\subsubsection{Impact of core sentence weight}

\begin{figure}[ht!]
    \centering
    \includegraphics[width=0.8\linewidth]{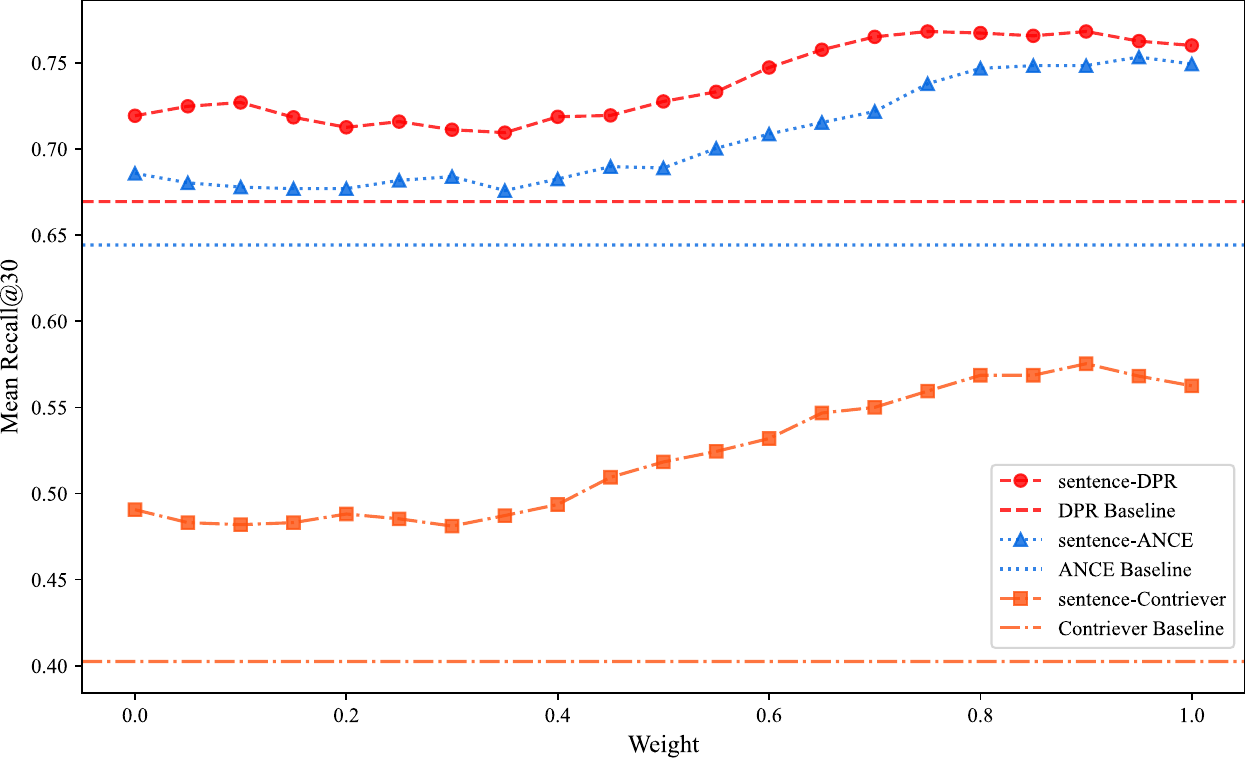}
    \caption{Impact of Core Sentence Weight on Recall across NQ Dataset.}
    \label{fig:the influence of core sentence weight between different retrievers}
\end{figure}

 From the Figure \ref{fig:the influence of core sentence weight between different retrievers}  it can be observed that when the weight of core sentences is adjusted to approximately 0.80, the Mean Recall@30 for ANCE, DPR and Contriever methods reaches optimal performance. This phenomenon reflects the impact of weight adjustment on the balance between contextual information and core sentences, which can be analyzed as follows:
 
\textbf{Performance Improvement at Optimal Weight (Around 0.80):} When the core sentence weight is set to approximately 0.80, the model effectively integrates contextual information with the content of core sentences. This balance enables the model to preserve semantic integrity while more accurately capturing key information relevant to the retrieval task, thereby achieving optimal recall performance.

\textbf{Performance Decline with Increased Weight (Beyond 0.80):} As the core sentence weight increases further toward 1.0, contextual information in the text is progressively diminished or even neglected, causing the model to rely more heavily on core sentences for retrieval. However, excessively weakening contextual information leads to a loss of semantic completeness, which adversely affects the accuracy of retrieval results. Consequently, performance declines beyond the 0.80 threshold.

\textbf{High Weight Still Outperforms the Baseline (At 1.0):} Even when the core sentence weight reaches 1.0, resulting in the complete disregard of contextual information, the model's performance remains superior to the baseline of paragraph-level retrieval. This indicates that paragraph-level information often contains significant redundancy, while core sentences play a pivotal role in enhancing retrieval performance. By adjusting the weighting, ParetoRAG effectively reduces the spatial burden of paragraph content while incorporating more core sentences, thereby improving retrieval precision and optimizing efficiency simultaneously.

\subsubsection{Impact of retriever} 
\label{sec:impact-of-retriever}
Figure \ref{fig:data-distribution} compares the ranking distribution of correct answers across different datasets (NQ, HotpotQA, and MS-MARCO) when using ParetoRAG and Naive RAG. The y-axis represents shows the percentage position of the correct answer within the ranked retrieval results, and the x-axis shows the fitted density distribution of the correct answer positions in the retrieval results. 20\% indicates that the correct answer appears in the top 20\% of the retrieval results. Higher percentage correspond to lower positions in the ranking, and values near -1 represent cases where the correct answer is not retrieved.


\begin{figure}[ht!]
    \centering
    \begin{subfigure}[b]{0.23\textwidth} 
        \centering
        \includegraphics[width=\linewidth]{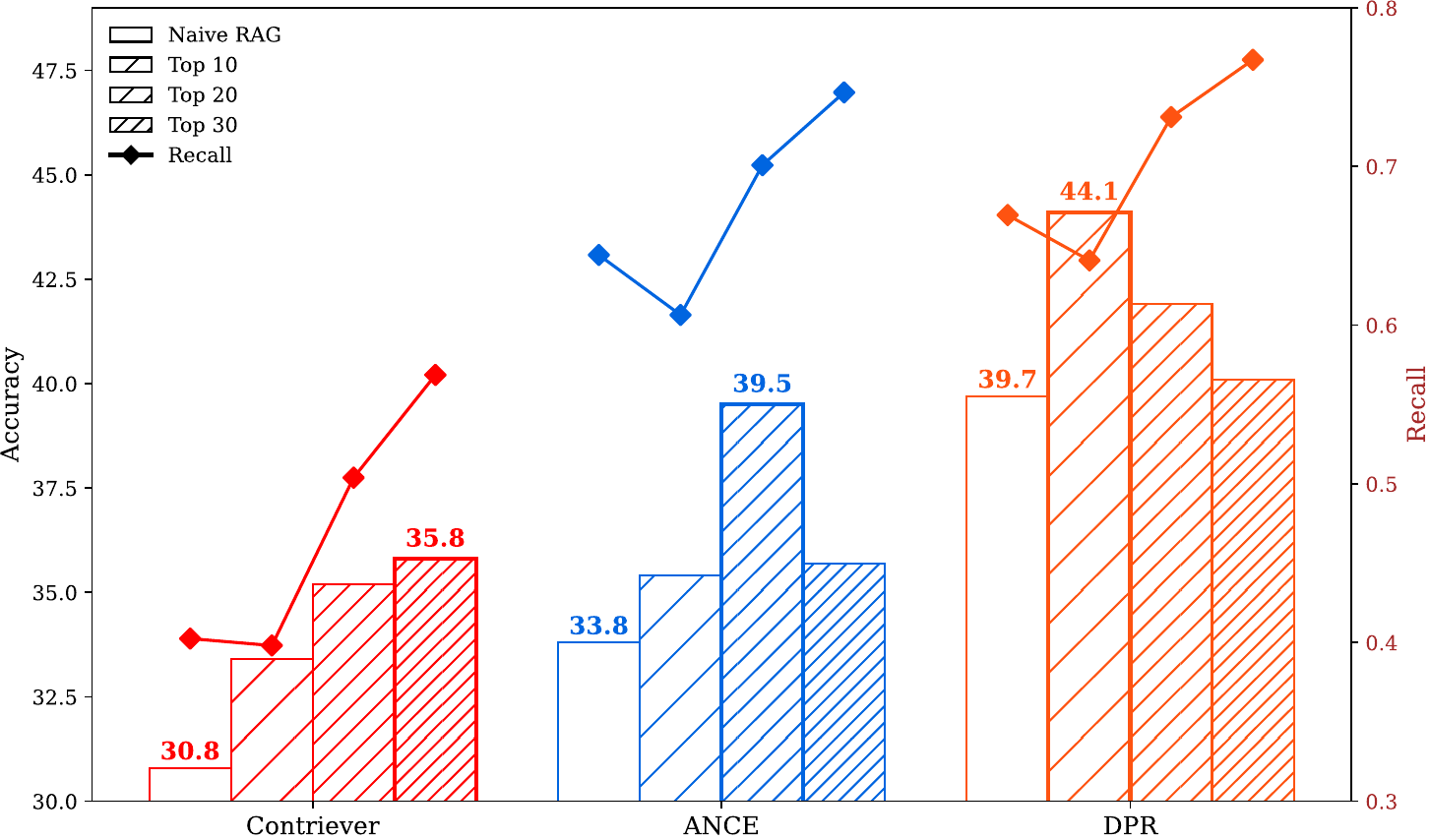}
        \caption{Vicuna-7B}
        \label{fig:vicuna-7b topk}
        \vspace{-5pt} 
    \end{subfigure}
    \begin{subfigure}[b]{0.23\textwidth} 
        \centering
        \includegraphics[width=\linewidth]{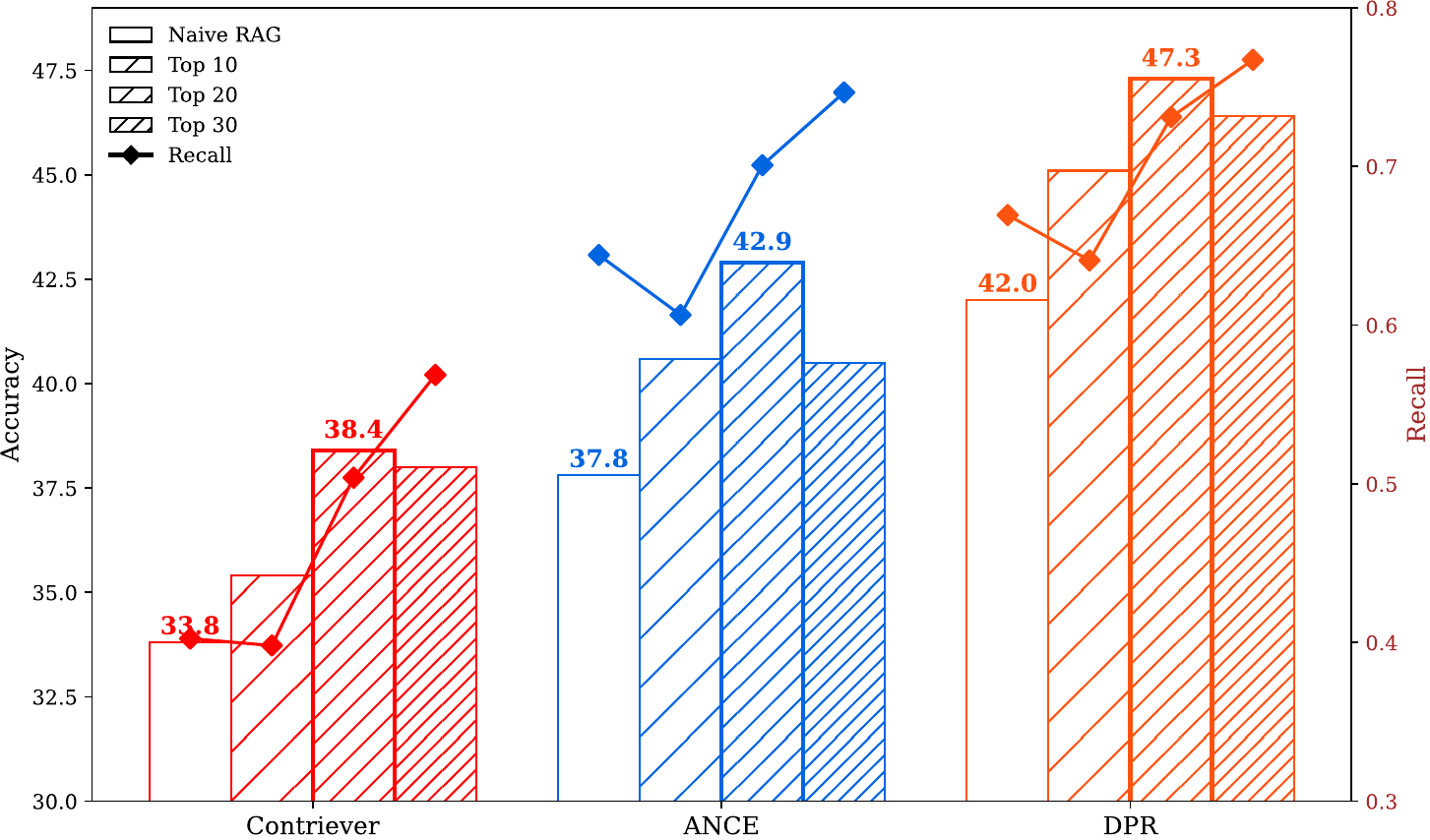}
        \caption{Vicuna-13B}
        \label{fig:vicuna-13b topk}
        \vspace{-5pt} 
    \end{subfigure}
    \caption{Comparison of accuracy and recall rates of different retrievers under various top k conditions.}
    \label{fig:impact of topk size}
\end{figure}

After being optimized by ParetoRAG, the three retrievers, DPR, ANCE, and Contriever, exhibit the following common trends: First, the correct answer rankings for all retrievers form a higher peak around 20\%, indicating that ParetoRAG effectively pushes correct answers to higher positions in the retrieval results. Second, the density near -1 is significantly reduced, demonstrating that ParetoRAG decreases the cases where correct answers are not retrieved, thus improving the retrieval comprehensiveness. 

Lastly, the distribution curves of ParetoRAG (blue lines) are smoother compared to Naive RAG (red lines), particularly in the mid-to-high ranking regions (e.g., 40\% to 80\%). This indicates that ParetoRAG stabilizes the performance of retrievers and reduces erroneous distributions.



\subsubsection{Impact of wider top k size}

Since the input size of ParetoRAG at Top 30 is similar to that of Naive RAG at Top 10 (more detailed can be seen in Appendix \ref{Appendix: Input Size Consistency}), we set the Top 10 performance of Naive RAG as the baseline. We then evaluate the performance of ParetoRAG in the top 10, top 20 and top 30 settings to investigate the impact of different top k configurations on model performance. As shown in Figure \ref{fig:impact of topk size}, our key observations are as follows:
\par
Although Naive RAG can achieve high document coverage, they often include a large amount of irrelevant information, which can interfere with the accuracy of LLM when answering questions. In contrast, with the fine-grained retrieval approach of ParetoRAG, although the recall rate is relatively lower under the same top k settings (e.g., Top 10), the accuracy of the language model's responses is significantly improved. This suggests that ParetoRAG, by performing document retrieval at the sentence level, effectively eliminates a substantial amount of irrelevant content, reducing LLM's inference complexity and enabling LLM to more efficiently identify the correct answer.
\par

 \subsubsection{Complementary effect of ParetoRAG on adaptive noise-robust LLM}
 \label{sec: detailed-analysis-on-robust-llm}
In this section, we evaluate the impact of ParetoRAG on robustly trained models, which are fine-tuned for the NQ and HotpotQA datasets respectively. These models are trained to enhance robustness against irrelevant context. As shown in Figure \ref{fig:ParetoRAGforRobustLLM}, in NQ, under the Top 10 retrieval setting, the accuracy improved from 44.80\% to 48.20\% (+3.4\%) when using ANCE. In HotpotQA, with input words length limited to 400 words, the accuracy increases from 23.3\% to 25.3\% (+2.0\%). These results demonstrate that ParetoRAG can further enhance performance in addition to robustly trained models.

While robust training improves the model's resilience to noisy contexts, it may still struggle with redundant or dense information in tasks involving long texts or multi-hop reasoning. ParetoRAG mitigates this limitation by reducing redundancy and increasing information density through sentence-level representations, allowing the model to focus more effectively on relevant content, thereby serving as a valuable complement to robustly trained models.

The complementary effect between ParetoRAG and robust training LLM indicates that combining these two approaches can further optimize retrieval and generation quality. Future work could explore integrating ParetoRAG with other training techniques to further enhance its performance across broader scenarios.


\section{Conclusion}
In this work, we propose ParetoRAG, an unsupervised framework that enhances RAG systems through sentence-level optimization guided by the Pareto principle. By decomposing paragraphs into sentences and dynamically re-weighting critical content while preserving contextual coherence, ParetoRAG achieves dual improvements in retrieval precision and generation quality without requiring additional training or API resources. Extensive experiments demonstrate its effectiveness: the framework reduces token consumption by 70\% while improving the accuracy and fluency of the answers in diverse datasets, LLMs and retrievers. Our analysis further reveals synergistic effects when integrating ParetoRAG with robustly trained language models, suggesting enhanced generalization capabilities. This study not only validates the viability of resource-efficient sentence-level refinement for RAG systems but also opens avenues for exploring hybrid methodologies that combine retrieval-augmented mechanisms with adaptive training strategies.

\section{Limitation}



While ParetoRAG demonstrates promising results in improving retrieval-augmented generation, it is important to acknowledge several potential limitations that could be addressed in future work. First, the sentence-level decomposition and re-weighting approach may weaken the complex cross-sentence logic or narrative connections within paragraphs, especially in tasks requiring multi-step reasoning or long-range semantic coherence (such as story generation or scientific argumentation). The local focus on key information might lead to a loose overall structure, which could impact the quality of the generated content. Second, when dealing with longer documents, ParetoRAG faces challenges related to segmenting the text and optimizing it at the sentence level. Breaking down long texts into sentences for individual optimization might not effectively preserve the global structure and logical flow of the document. Lastly, while ParetoRAG has been tested on open-domain QA datasets, it has not yet been applied to more specialized domains, such as law or medicine, which could be explored in future work.
\bibliography{acl_latex}

\appendix

\section{System Prompt} \label{appendix:prompt}


The following is the system prompt used to let a LLM generate an answer without any information:

\begin{mdframed}[
    backgroundcolor=gray!10, 
    linecolor=black, 
    linewidth=0.5pt, 
    roundcorner=5pt, 
    innertopmargin=10pt, 
    innerbottommargin=10pt, 
    innerleftmargin=10pt, 
    innerrightmargin=10pt, 
    skipabove=10pt, 
    skipbelow=10pt 
]
\setlength{\parskip}{2pt}
You are a helpful assistant. 
Answer the question as concisely as possible, using only the specific phrase, entity, or number that directly answers the question. Within five words. \\
\noindent \textbf{Query: }[question]  \\
\noindent \textbf{Short Answer:}
\end{mdframed}

The following is the system prompt used in RAG to let a LLM generate a NQ answer based on the given context:

\begin{mdframed}[
    backgroundcolor=gray!10, 
    linecolor=black, 
    linewidth=0.5pt, 
    roundcorner=5pt, 
    innertopmargin=10pt, 
    innerbottommargin=10pt, 
    innerleftmargin=10pt, 
    innerrightmargin=10pt, 
    skipabove=10pt, 
    skipbelow=10pt 
]
\setlength{\parskip}{2pt}
You are a knowledgeable assistant tasked with answering questions based on the Natural Questions dataset. 
Each question is accompanied by contexts extracted from Wikipedia. 
Answer the question by providing only the specific phrase, entity, or number that directly answers the question. Within five words.

\noindent\textbf{Contexts:} [context]

\noindent\textbf{Query:} [question]

\noindent\textbf{Short Answer:}
\end{mdframed}

The following is the system prompt used in RAG to let a LLM generate a MS answer based on the given context:

\begin{mdframed}[
    backgroundcolor=gray!10, 
    linecolor=black, 
    linewidth=0.5pt, 
    roundcorner=5pt, 
    innertopmargin=10pt, 
    innerbottommargin=10pt, 
    innerleftmargin=10pt, 
    innerrightmargin=10pt, 
    skipabove=10pt, 
    skipbelow=10pt 
]
\setlength{\parskip}{2pt}
You are a knowledgeable assistant tasked with answering questions based on the MS-marco dataset. 
Answer the question given the information in those contexts.
Answer the question in a single, brief sentence.

\noindent\textbf{Contexts:} [context]

\noindent\textbf{Query:} [question]

\noindent\textbf{Answer:}
\end{mdframed}

The following is the system prompt used in RAG to let a LLM generate a HotpotQA based on the given context:

\begin{mdframed}[
    backgroundcolor=gray!10, 
    linecolor=black, 
    linewidth=0.5pt, 
    roundcorner=5pt, 
    innertopmargin=10pt, 
    innerbottommargin=10pt, 
    innerleftmargin=10pt, 
    innerrightmargin=10pt, 
    skipabove=10pt, 
    skipbelow=10pt 
]
\setlength{\parskip}{2pt}
You are a knowledgeable assistant tasked with answering questions based on the HotPotQA dataset. 
Each question is accompanied by contexts extracted from Wikipedia. 
Answer the question as concisely as possible, using only the specific phrase, entity, or number that directly answers the question. Within five words.

\noindent\textbf{Contexts:} [context]

\noindent\textbf{Query:} [question]

\noindent\textbf{Short Answer:}
\end{mdframed}

\section{Calculation of Document Retrieval Ratio for Input Size Consistency} \label{Appendix: Input Size Consistency}

In the original corpus, the average token count per paragraph for the top 30 retrieved paragraphs is 80.8575 tokens. After applying the Sentence-RAG, the average token count per paragraph for the top 30 retrieved paragraphs decreases to 23.85 tokens. Therefore, to maintain consistency in the total token count of input paragraphs, Sentence-RAG would theoretically need to retrieve the top $80.8575/23.85 \approx 34$ paragraphs to match the token scale of the top 10 paragraphs retrieved by naive-RAG. While strict calculations suggest retrieving approximately 34 paragraphs, selecting 30 paragraphs strikes a balance between maintaining the validity of experimental results and ensuring clarity and simplicity in presentation.

\section{Statistics of Datasets}
\begin{table}[!ht]
    \centering
    \resizebox{0.45\textwidth}{!}{
    \begin{tabular}{cccc}
    \hline
        ~ & NQ & HotpotQA & MS-MARCO  \\ \hline
        Passages & 2,681,468 & 5,233,329 & 8,841,823  \\ 
        Ours & 9,320,506 & 12,425,366 & 30,137,968  \\ \hline
    \end{tabular}
    }
    \label{table:statistic-dataset}
    \caption{Statistics of text unit counts before and after ParetoRAG encoding. }
    \label{table:statistic-dataset}
\end{table}

\begin{figure*}[ht!]
    \centering
    \begin{subfigure}[b]{0.48\textwidth} 
        \centering
        \includegraphics[width=\linewidth]{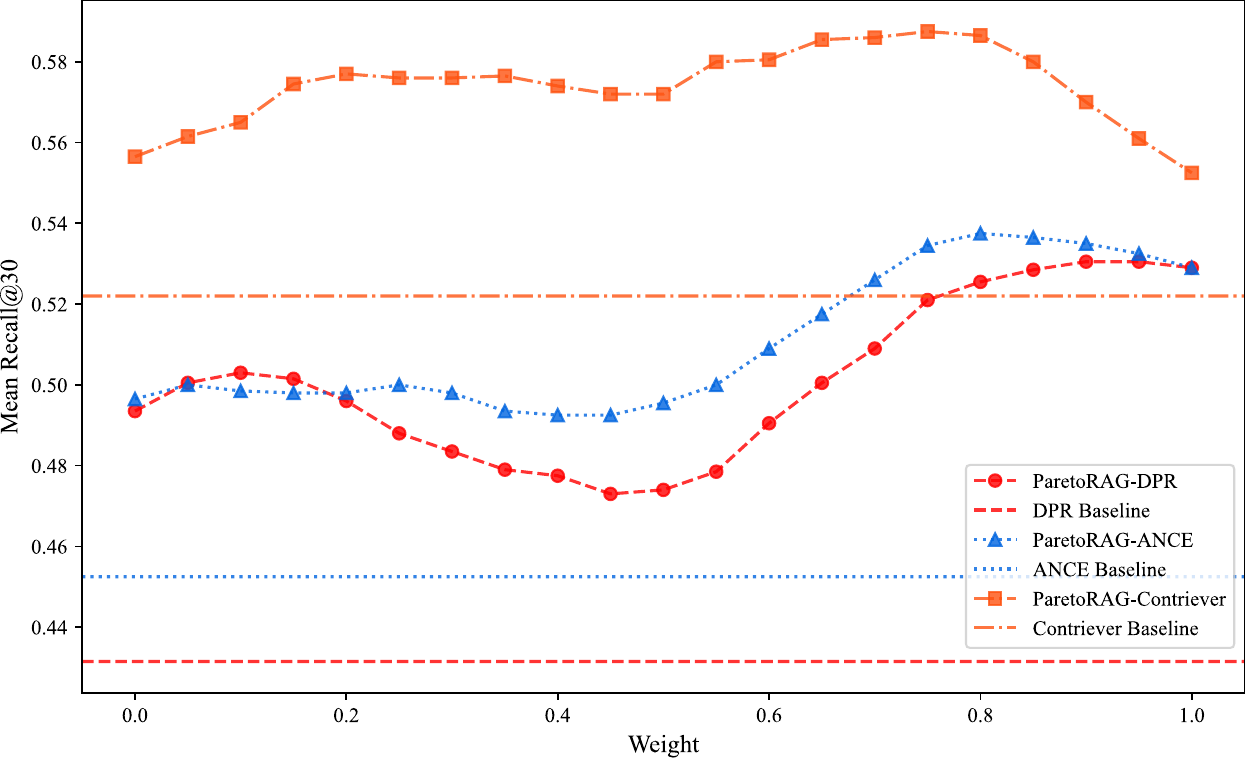}
        \caption{HotpotQA}
        \label{fig:HotpotQA Core Sentence}
        \vspace{-5pt} 
    \end{subfigure}
    \begin{subfigure}[b]{0.48\textwidth} 
        \centering
        \includegraphics[width=\linewidth]{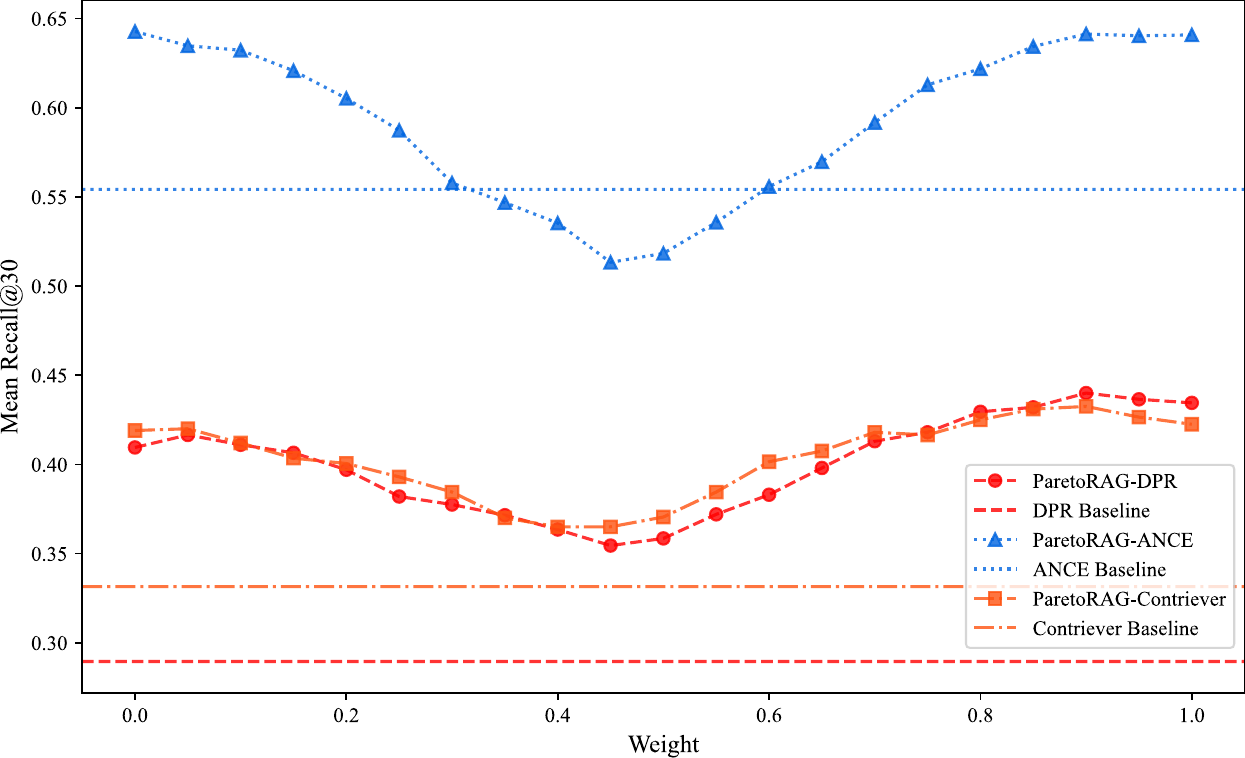}
        \caption{MS-MARCO}
        \label{fig:MS-Marco Core Sentence}
        \vspace{-5pt} 
    \end{subfigure}
    \caption{Impact of Core Sentence Weight on Recall across HotpotQA and MS-Marco Dataset. }
    \label{fig:impact of topk size appendix}
\end{figure*}

Table \ref{table:statistic-dataset} shows statistics of text unit counts before and after ParetoRAG encoding.

\section{Statistics of Models} \label{appendix: Statistics of models}
All model weights are derived from Hugging Face, which were used without additional training. In the following, we list the specific hugging face model names corresponding to the weights used in the experiment:

\subsection{Model Weights}
\begin{itemize}[leftmargin=*]
    \small
    \item \textbf{DPR}:
    \begin{itemize}
        \item \path{facebook/dpr-question_encoder-multiset-base}
        \item \path{facebook/dpr-ctx_encoder-multiset-base}
    \end{itemize}
    \item \textbf{Contriever}:
    \begin{itemize}
        \item \path{facebook/contriever}
    \end{itemize}
    \item \textbf{ANCE}:
    \begin{itemize}
        \item \path{sentence-transformers/msmarco-roberta-base-ance-firstp}
    \end{itemize}
    \item \textbf{RECOMP}:
    \begin{itemize}
        \item \path{fangyuan/nq_abstractive_compressor}
        \item \path{fangyuan/hotpotqa_abstractive_compressor}
    \end{itemize}
    \item \textbf{Llama2}:
    \begin{itemize}
        \item \path{meta-llama/Llama-2-7b-chat-hf}
        \item \path{meta-llama/Llama-2-13b-chat-hf}
    \end{itemize}
    \item \textbf{Viccuna}:
    \begin{itemize}
        \item \path{lmsys/vicuna-7b-v1.3}               
        \item \path{lmsys/vicuna-13b-v1.3}
    \end{itemize}
    \item \textbf{RetRobust}:
    \begin{itemize}
        \item \path{Ori/llama-2-13b-peft-nq-retrobust}               
        \item \path{Ori/llama-2-13b-peft-hotpotqa-retrobust}
    \end{itemize}

\end{itemize}

\subsection{Model Hyperparameter} 
The model's configuration is as follows: \path{max_output_tokens} is set to 150, limiting the maximum number of tokens in the generated output; \path{temperature} is set to 0.1, which controls the randomness of the generation process, ensuring more deterministic and focused outputs; \path{seed} is fixed at 100 to ensure reproducibility of the results across different runs; and \path{per_gpu_batch_size} is set to 16, specifying the number of samples processed per GPU in each batch during training or inference.

\section{Evaluation Metrics} \label{appendix: Evaluation Metrics}
Following the experimental setup in \cite{asaiSelfRAGLearningRetrieve2023a}, we use MAUVE\cite{pillutlaMAUVEMeasuringGap2021} and ROUGE-L \cite{linROUGEPackageAutomatic2004} as evaluation metrics for long-form generation. For short-form generation, we use accuracy (ACC). For each question, if the standard answer is contained within the generated answer and the length of the generated answer is less than or equal to 15, it is counted as 1. Here’s a brief explanation of the evaluation metrics: 
\begin{itemize}
    \item Accuracy: Measures the percentage of correct predictions made by the model. It's a basic metric that indicates how well the model is performing on a classification or question-answering task.
    \item ROUGE: Evaluates text summarization or generation by comparing the overlap between generated text and reference text. It focuses on recall, ensuring the generated text captures key information from the reference. Common variants include ROUGE-N (n-gram overlap) and ROUGE-L (longest common subsequence).
    \item MAUVE:  Assesses text generation quality by comparing the distribution of generated text to reference text in an embedding space. It uses divergence measures to evaluate semantic and structural alignment, making it particularly useful for open-ended tasks like story or dialogue generation.
\end{itemize}

\section{Impact of core sentence weight} \label{appendix: Impact of core sentence weight}

As shown in the Figure \ref{fig:impact of topk size appendix}, HotpotQA and MS-MARCO generally follow the trends analyzed in Section 5.2.1, where increasing the core sentence weight typically improves recall performance. However, there are noticeable differences in the details of recall variations between these two datasets. Specifically, in the MS-MARCO dataset, the recall rates of ParetoRAG-DPR and ParetoRAG-ANCE decrease significantly in the core sentence weight range of 0.3 to 0.6. This phenomenon can be attributed to the following key factors:

\subsection{Differences in Retrieval Model Training Approaches}
\begin{itemize}
    \item \textbf{Asynchronous Global Index Updates versus Local Contrastive Learning}: ANCE relies on \textit{asynchronous global index updates}, whereas DPR and Contriever adopt \textit{local contrastive learning} with positive and negative samples. This distinction makes ANCE more susceptible to contextual noise when the core sentence and contextual sentence weights are close (0.3–0.6), leading to less precise retrieval and subsequently lower recall performance.
    \item In contrast, DPR and Contriever primarily depend on \textit{local contrastive learning} during training. Since they do not suffer from the lag introduced by global index updates, their recall rate decline in the 0.3–0.6 weight range is relatively less pronounced.
\end{itemize}

\subsection{Differences in Task Types and Information Requirements}
\begin{itemize}
    \item \textbf{MS-MARCO (Single-hop QA)}: In this dataset, queries typically require matching a specific \textit{core sentence} in the text to retrieve the correct answer, while paragraph-level information may contain substantial redundant content. Consequently, when the core sentence weight falls within the 0.3–0.6 range, paragraph-level information introduces interference in the retrieval process, leading to a decline in recall performance.
    \item \textbf{HotpotQA (Multi-hop QA)}: In contrast, HotpotQA involves multi-hop reasoning, where queries require integrating information from multiple paragraphs to derive the final answer. As a result, even when the core sentence weight is relatively low, the model can still leverage other paragraphs to improve retrieval performance. Therefore, unlike MS-MARCO, HotpotQA does not exhibit a sharp decline in recall within the 0.3–0.6 weight range.
\end{itemize}

\section{Impact of Model Size on Accuracy with Varying Top K in ParetoRAG} \label{appendix: Impact of Model Size on Accuracy with Varying Top K in ParetoRAG}

As show in Figure \ref{fig:impact of topk size}, for smaller models (such as Vicuna-7b), their ability to process a large number of documents is weaker, leading to a faster decline in accuracy as the Top K increases. However, this decline is not due to a lack of retrieval quality by ParetoRAG, but rather because smaller models are unable to fully utilize the richer information provided. On the other hand, for larger models (such as Vicuna-13b), their greater parameter size and reasoning capability enable them to handle more information within a larger scope. As a result, even when the Top K is increased to a certain extent (e.g., Top 20 or Top 30), they still maintain high accuracy. Notably, larger Top K settings (e.g., Top 20) outperform Top 10 and the baseline, demonstrating that ParetoRAG can provide richer information retrieval, offering more effective context for language models.

\end{document}